\documentclass[preprint,12pt]{cas-sc}

\def\tsc#1{\csdef{#1}{\textsc{\lowercase{#1}}\xspace}}
\tsc{WGM}
\tsc{QE}

\usepackage{times}
\usepackage{epsfig}
\usepackage{graphicx}
\usepackage{amsmath}
\usepackage{amssymb}
\usepackage{xcolor, color, soul}
\sethlcolor{red}

\usepackage{balance}
\usepackage{array,multirow}
\usepackage{hyperref}
\usepackage[numbers]{natbib}

\usepackage{color}
\definecolor{red}{rgb}{1.00,0.00,0.00}
\definecolor{blue}{rgb}{0.00,0.00,1.00}
\definecolor{green}{rgb}{0.3,0.75,0.0}
\definecolor{yellow}{rgb}{0.5,0.5,0.0}
\definecolor{orange}{rgb}{1.,0.5,0.0}

\usepackage{subcaption}

\begin{document}
\let\WriteBookmarks\relax
\def\floatpagepagefraction{1}
\def\textpagefraction{.001}



\title [mode = title]{Dense Extreme Inception Network for Edge Detection} 
\shorttitle{Dense Extreme Inception Network for Edge Detection}

\shortauthors{X. Soria et.al.} 
%

\author[1,2]{Xavier Soria}[
      orcid=0000-0003-2997-2439]

\cormark[1]
\ead{xavier.soria@unach.edu.ec}

\author[3,1]{Angel Sappa}[
      orcid=0000-0003-2468-0031,]
\ead{sappa@ieee.org}

\author[2]{Patricio Humanante}[
      orcid=0000-0003-2632-2051]
\ead{phumanante@unach.edu.ec}

\author[4]{Arash Akbarinia}[
      orcid=0000-0002-4249-231X,]
\ead{Arash.Akbarinia@psychol.uni-giessen.de}

\affiliation[1]{organization={Computer Vision Center, Autonomous University of Barcelona},
            addressline={}, 
            city={Barcelona},
            country={Spain}}

\affiliation[2]{organization={UMAYUK Research Group, National University of Chimborazo},
            city={Riobamba},
            country={Ecuador}}
            
\affiliation[3]{organization={ESPOL Polytechnic University, FIEC, CIDIS},
            city={Guayaquil},
            country={Ecuador}}            

\affiliation[4]{organization={Department of Experimental Psychology, Justus-Liebig University},
            city={Giessen},
            country={Germany}}
            

\cortext[1]{Corresponding author}
\fntext[1]{Xavier Soria}



\begin{abstract}
$\lll$This is a $pre-acceptance$ $version$, please, go through Pattern Recognition Journal on Sciencedirect to read the $final$ $version$$\ggg$. Edge detection is the basis of many computer vision applications. State of the art predominantly relies on deep learning with two decisive factors: dataset content and network's architecture. Most of the publicly available datasets are not curated for edge detection tasks. Here, we offer a solution to this constraint. First, we argue that edges, contours and boundaries, despite their overlaps, are three distinct visual features requiring separate benchmark datasets. To this end, we present a new dataset of edges. Second, we propose a novel architecture, termed Dense Extreme Inception Network for Edge Detection (DexiNed), that can be trained from scratch without any pre-trained weights. DexiNed outperforms other algorithms in the presented dataset. It also generalizes well to other datasets without any fine-tuning. The higher quality of DexiNed is also perceptually evident thanks to the sharper and finer edges it outputs.
\end{abstract}




\maketitle
\tnotetext[]{}
\section{Introduction}
\label{sec:intro}
\begin{figure*}
\begin{center}
    \includegraphics[width=0.87\textwidth,height=0.76\textheight]{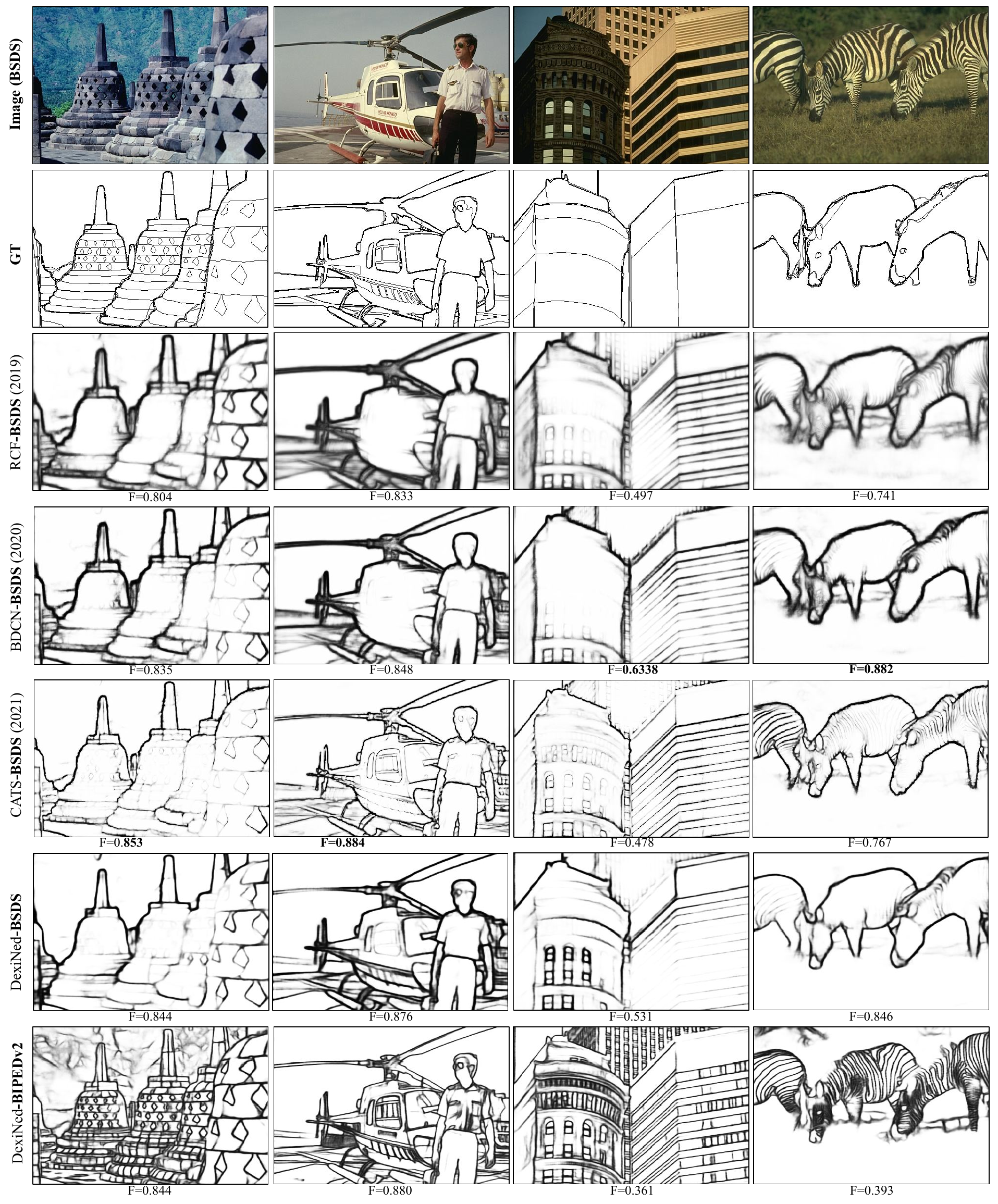}
\end{center}
   \caption{The results of several algorithms on the BSDS dataset \cite{arbelaez2011bsds500}. The suffix in model names (i.e., BSDS and BIPEDv2) indicates the training set of that model.}
\label{fig:banner}
\end{figure*}

Edges provide important clue in visual information processing, from classical computer vision algorithms in image recognition \cite{nmakram2008} to modern techniques in generative adversarial networks (GAN) \cite{zhu2017cyclegan}. Despite their importance, robust edge detection remains an open problem. To demonstrate this, let's examine four example images in Fig. \ref{fig:banner}. The third to fifth rows illustrate the results of three state-of-the-art models (RCF \cite{liu2019RCFext}, BDCN \cite{he2020bdcnPAMI}, and CATS \cite{huan2021cats}) that are trained on the BSDS dataset \cite{arbelaez2011bsds500}. It is qualitatively visible that neither of these models faithfully detects the edges of tiny details. For instance, while the internal edges of the bell sculpture or the helicopter are fully annotated in the ground truth, the edge output of those models do not capture these details. We can observe a similar phenomenon in the stripes of the building and the zebra. In this case, although these details are also excluded in the ground truth, the edges are clearly visible in the image. Grounded on this, we argue that current models of edge detection require further improvement to robustly detect edges and generalize well to new scenes independent of their training set.

Following this, we argue that one of the main limitations of current deep learning (DL) approaches is the annotated ground truth in the training set. BSDS is widely accepted within the community as the benchmark dataset to train and evaluate edge-detection algorithms. However, this dataset was originally designed and annotated for scene segmentation. Therefore the annotated ground truth correspond mainly to high-level object boundaries rather than low-level edges. Given the DL based models are strongly shaped by their training data, we need to acquire independent datasets for three tasks of edge-, contour, and boundary-detection \cite{mely2016multicue}. This is of great importance for both training and evaluation. It is challenging for a network to learn these three concepts from a dataset that blends edges, contours and boundaries. It is also difficult to judge the fitness of a network, whether it is performing better or worse in one of those tasks.


To showcase this importance at a glance, we can look at the bottom two rows of Fig. \ref{fig:banner}. The only difference between those two rows is the training set. When our network---Dense Extreme Inception Network for Edge Detection (DexiNed)---is trained on BSDS, it suffers from the same set of problems as other models. However, when we train DexiNed on our dataset (BIPEDv2), it generalizes well to BSDS images in fine-scale edges. The image content of these two datasets is rather dissimilar. The BIPEDv2 mainly contains images of the urban settings. The accurate detection of edges in the zebra and bell sculpture images suggests that the DexiNed-BIPEDv2 is robust to novel scenes. This robustness is thanks to the careful annotation of edges in the proposed dataset. We ensured all the edges, within or across objects, are reflected in the ground truth data.



\subsection{Edges, contours and boundaries}
The following three terms: edge-, contour- and boundary-detection are often used interchangeably despite their differences. This is a potential cause of confusion in the interpretation and evaluation of models. Thus, we start by reviewing the origin of each term. In the 1980s, edge detection was defined as the intensity changes in a vicinity originated by discontinuities along the surface, reflectance, or illumination \cite{canny1987cannymethod}. This was revisited emphasizing the properties of objects, such as their photometrical, geometrical and physical characteristics  \cite{ziou1998edgeOverview}. Accordingly, contours and boundaries are a subset of edges and are associated with semantically meaningful entities \cite{Gong2018contourOverview}, for example, the silhouette and outline of an object \cite{ming2016contourDef}. Boundary refers to the object borders in the image plane, corresponding to pixel ownership within a scene \cite{grigorescu2003cid, martin2004mBSDS300ext}. On the contrary, contour refers to the borders of a region within a given object \cite{mely2016multicue}.

To summarize, the goal of an edge detection algorithm is to capture all the meaningful intensity discontinuities in an image. It does not concern itself with pixel ownership or open and closed shapes. These are the domain of contour and boundary detection to eliminate edges that do not correspond to salient features of objects and shapes. In this article, we focus on edges.


\begin{figure*}
\centering
\begin{tabular}{c}
     \includegraphics[width=0.75\textwidth]{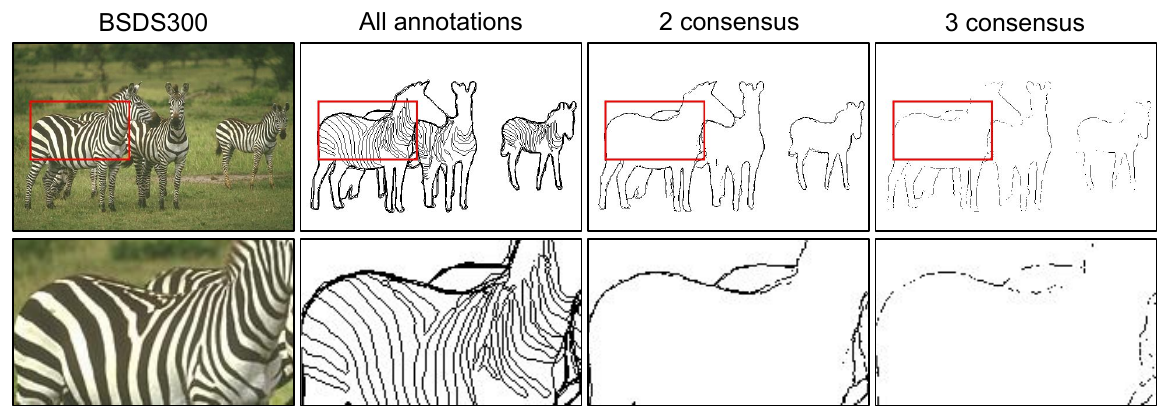} \\
     (a) BSDS300 \cite{martin2004mBSDS300ext}\\
\includegraphics[width=0.75\textwidth]{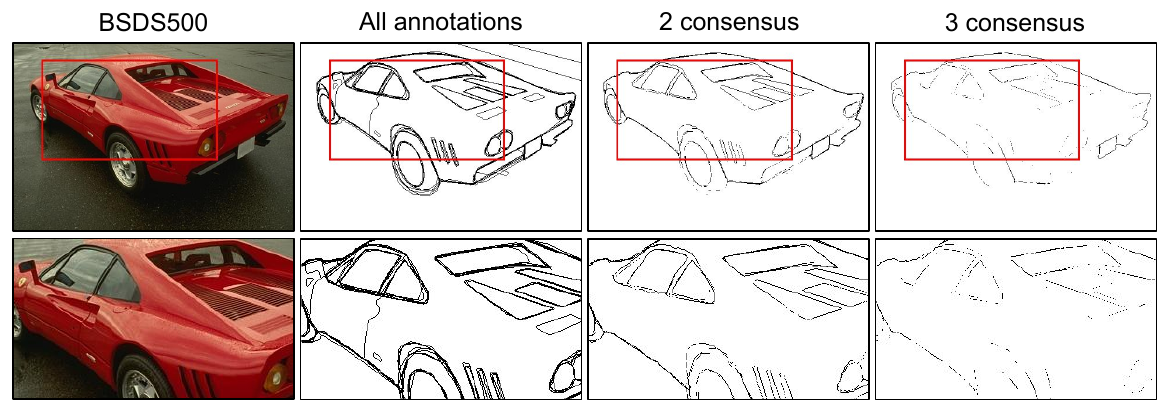}\\
(b) BSDS500 \cite{arbelaez2011bsds500}\\
\end{tabular}
\caption{A careful examination of the the BSDS benchmark datasets for two sample images and their corresponding ground truths. There is a large discrepancy among human-drawn ``edges''. Enforcing consensus among two/three annotations results in the elimination of true edges. This issue demonstrates that BSDS is a suitable dataset to evaluate boundaries but \underline{\textbf{not}} edges.}
\label{fig:gt-errs}
\end{figure*}

\subsection{Structure of data matters}

Currently, deep learning is the primary approach in the task of edge detection. It is well established that the quality of the training dataset is a critical factor within this framework \cite{lecun2015deep}. To this end, state-of-the-art heavily relies on the Berkeley segmentation dataset (i.e., BSDS300 \cite{martin2004mBSDS300ext} and BSDS500 \cite{arbelaez2011bsds500}) even though image segmentation was the original objective of this dataset and its refined version turned into a benchmark dataset of boundary detection. Most studies train their models on BSDS300 and validate on BSDS500. 

Each image contains boundary annotations drawn by at least five different participants. In many cases, there is a large discrepancy among different human-drawn boundaries (see Fig. \ref{fig:gt-errs}). This issue is partially overcome by computing the loss function over a consensus ground truth, e.g., \cite{xie2017hed,liu2019RCFext, he2020bdcnPAMI}. In the consensus stage, pixels that are annotated by more than two participants are considered as true edges, otherwise discarded. This facilitates the convergence of the training process \cite{xie2017hed}. Nevertheless, this consensus stage also filters out a large portion of true edges. This is visually evident in Fig. \ref{fig:gt-errs}. For instance, all the zebra stripes vanish even with a moderate consensus of two participants. A similar observation can be made in the case of the red car. This problem exists in many images of the BSDS.

Hence, the networks trained on the BSDS learn to become a \textit{boundary detector} rather than an \textit{edge detector}. The distinction between the two can be of great importance, for example, when edges are the building blocks of another computer vision application, as in \cite{pourreza2017medImg}. This issue is addressed in this study. We present a benchmark dataset of edges that allows for an accurate evaluation of edge detection algorithms.


\subsection{Contributions}

In this paper, we address the aforementioned issues by proposing an end-to-end deep learning approach for the task of edge detection. Our primary objective is to identify all true edges in any given image. To demonstrate this, we train a network only on one dataset and evaluate it on several publicly available datasets. We show that our model obtains a higher degree of generalization in comparison to the other algorithms of state-of-the-art. We conduct an exhaustive quantitative evaluation on two datasets with annotated edges (i.e., MDBD \cite{mely2016multicue} and BIPED). On other datasets with contour/boundary annotation, we qualitatively show the robustness of our model.
In the current work, we extend our previous conference paper \cite{soria2020dexined} in the following aspects:
\begin{itemize}
    \item Presenting a detailed description of the proposed architecture, DexiNed (Dense eXtreme Inception Network for Edge Detection), and thoroughly analysing the impact of its different components.

    \item Boosting the degree of annotation, specifically for fine scale edges, in the proposed benchmark dataset of edge detection, referred to as BIPED (Barcelona Images for Perceptual Edge Detection)\footnote{Dataset and code: \url{https://github.com/xavysp/DexiNed}}.
    
    
    \item Establishing a common evaluation benchmark among four state-of-the-art networks by interchanging the training and validation set between two datasets of edge detection (BIPED and MDBD).
    
    \item Improving the loss functions from BDCN \cite{he2020bdcnPAMI} by modifying the $\lambda$ values to better balance the portion of positive and negative samples in each DexiNed output and incorporating averaging at the level of pixels.  
\end{itemize}

The rest of the paper is organized as follow. Section~\ref{sec:rw} summarizes the most relevant  works on edge detection. Then, the proposed approach is described in Section \ref{sec:pa}. Acquired dataset and ground truth generated for training and testing, as well as the datasets used for validating the proposed DexiNed architecture, are presented in Section \ref{sec:data}. In Section \ref{sec:res}, quantitative and qualitative details are summarized. Finally, conclusions are given in Section~\ref{sec:con}.


\section{Related Work}\label{sec:rw}

There is a large body of literature on edge detection, for a detailed description see reviews on \cite{ziou1998edgeOverview} and \cite{Gong2018contourOverview}. In this section, a set of representative algorithms are detailed. They can be broadly categorized into four groups: $i)$ driven by low-level features; $ii)$ brain-inspired; $iii)$ classical learning-based; and $iv)$ deep learning.

\textit{Algorithms driven by low-level features}: The early edge detection algorithms such as Sobel \cite{sobel1972sobelmethod}, Robert and Prewitt \cite{Rosenfeld1982firstDeriv} are based on the first-order derivative. The input image is smoothed by the linear local filters, normally, a set of two orientations (horizontal and vertical). In the end, edges are detected by thresholding. These operators advanced by considering the second-order derivative, where edges are detected by the extraction of zero-crossing points. In the mid-eighties Canny \cite{canny1987cannymethod} proposed an edge detector by grouping three different key processes.
Despite being antiquated, these approaches are still used in some modern computer vision applications. Many variants of the above-mentioned algorithms have been present in the literature. 

\textit{Brain-inspired algorithms:} This set of edge detection algorithms rely on known mechanisms of the biological visual system. Since the 1960s, experiments on monkeys and cats started resulted in important advances on the understanding of Primary Visual Cortex (V1). For instance, it was discovered that a group of simple cells are responsive to edges \cite{young1987BIMinspir}. The authors in \cite{young1987BIMinspir} develop a mathematical model to simulate the human retinal vision by Gaussian derivatives; since then, many algorithms have been implemented resembling the edge detection of the biological neurons in real-world images. For example, in \cite{basu1994mBIM0} a special line weight function is introduced for edge enhancement, which consists of a combination of zero and second-order Hermite functions. Later on, in \cite{grigorescu2003cid} Gabor energy maps have been proposed to recreate the non-classical receptive field of V1 for contour detection. This proposal has been evaluated with 40 images---this dataset could be considered as the first annotated ground truth proposed in the literature to validate contour detection algorithms. This approach has been later on improved by \cite{tang2007mBI3} by modeling spatial facilitation and surround inhibition with local grouping functions and Gabor filters respectively. Recently, in \cite{yang2015SCO}, a modulation of double opponent cells is performed to extract more complex edge properties from color and texture. This was complemented by the introduction of the second visual area and accounting for feedback connections \cite{akbarinia2018feedback}. Subsequent experiments demonstrated that modeling various surround modulations, typical in neurons of the visual cortex, boost the performance of edge detection \cite{akbarinia2018feedback}. A combination of three Gabor filters at different scales was proposed in \cite{mely2016multicue} followed by a PCA and a machine-learning classifier. Lading us to the next group that are learning-based.

\textit{Classical learning-based algorithms:} The challenge of edge detection in natural scenes has motivated the development of the learning-based algorithm. One of the first learning approaches for boundary detection has been presented in \cite{martin2004mBSDS300ext}, which uses different filters to extract gradients from brightness, color and texture. The authors then train a logistic regression classifier to generate a Probabilistic boundary (Pb). Later on, different variants of Pb detectors have been proposed, like the one from \cite{maire2008mgPb1}. These new proposals focus on global information besides using local information like the original approach. They are referred to as globalized Probability of boundary (gPb). In these approaches, the gradients are computed by three scales for each image channel individually. In \cite{ren2005mCLA2}, a conditional random field-based approach has been proposed. Its maximum likelihood parameters are trained given the image edge annotations. This model captures the continuity and frequency of different junctions by a continuity structure. On the contrary, the usage of a sparse code gradient has been proposed in \cite{xiaofeng2012mCLAsCode}. In this case, the patches are classified by a support vector machine. Lastly, the usage of a Bayesian model has been considered in \cite{widynski2014mPb2ext}. Edge-maps have been predicted by a Sequential Monte Carlo based approach using different gradient distributions. The random forest framework is also considered for edge detection, where each tree is trained independently to capture complex local edge structures \cite{dollar2015forests}.

\textit{Deep learning algorithms:} During the last decade, CNNs have become the standard model in computer vision \cite{krizhevsky2012alexnet}. It has been shown that the internal representation of object classification networks relies on edges \cite{akbarinia2020deciphering}. The first CNN based model for edge detection has been proposed in \cite{ganin2014firstDLedge}. In this approach, like in most DL models, given an annotated edge-map dataset, the network learns to predict edges from an RGB image. Since this original work, several methods have been proposed \cite{Gong2018contourOverview}; the backbone of most architectures is VGG16 \cite{simonyan2014vgg}. From this architecture, just the convolutional layers are used (13 convolutional layers). The parameters of this architecture are obtained by training the network in the ImageNet dataset \cite{krizhevsky2012alexnet} as a classification problem. Then, the network is fine-tuned by training it on the datasets conceived for boundary detection. This is the case of HED \cite{xie2017hed}, RCF \cite{liu2019RCFext}, CED \cite{wang2017ced}, BDCN \cite{he2020bdcnPAMI}, CATS \cite{huan2021cats} and many others. In HED \cite{xie2017hed} a multi-scale network together with a deep supervision technique is configured. Hence, each VGG block has as an output an edge prediction, which can be considered as a space scale representation. RCF \cite{liu2019RCFext} uses the same configuration as HED, but instead of getting edges from each VGG block, it extracts edges from each layer on the blocks.  On the contrary to previous approaches, in CED refinement blocks are used to merge outputs from every single block and predicts boundaries sharper than the ones from HED. Even though the quantitative performance of the aforementioned methods overcomes the state-of-the-art of classical learning algorithms, the predicted edge-maps are not as sharp as expected, somehow detected edges are coarse. Hence, Generative Adversarial Networks (GAN) \cite{goodfellow2014gan} have been considered to sharpen those coarse edges (e.g., \cite{yang2019contourgan}); lastly, Context-Aware Tracing Strategy (CATS) is proposed to improve the edge-map crispness.

Although most of the DL based models for edge detection have been based on the usage of VGG16, there are a few approaches that are based on other models, like ResNet50 \cite{he2016resnet} that is used in \cite{liu2019RCFext}. With these variants, tinny improvements have been obtained, about $1\%$. Despite that, in the current work VGG16 is going to be described as the standard architecture used on the edge detection approaches. Most of the models based on VGG16 outperform traditional edge detection methods in the standard edge detection datasets such as BSDS \cite{arbelaez2011bsds500}, NYUD \cite{silberman2012NYUD}, PASCAL-CONTEXT \cite{mottaghi2014PASCALcontext} and MDBD--- Multicue Dataset for Boundary Detection \cite{mely2016multicue}. Despite this, even after obtaining high ranked performance, some of these methods still present drawbacks or lack of generalization or coarser predicted edge-maps, see results in Fig. \ref{fig:banner}. For instance, it can be observed that in some cases the generated edge-maps do not predic the human perceptual edges.

\begin{figure}
\centering
\includegraphics[width=0.75\textwidth,height=0.8
\textheight]
{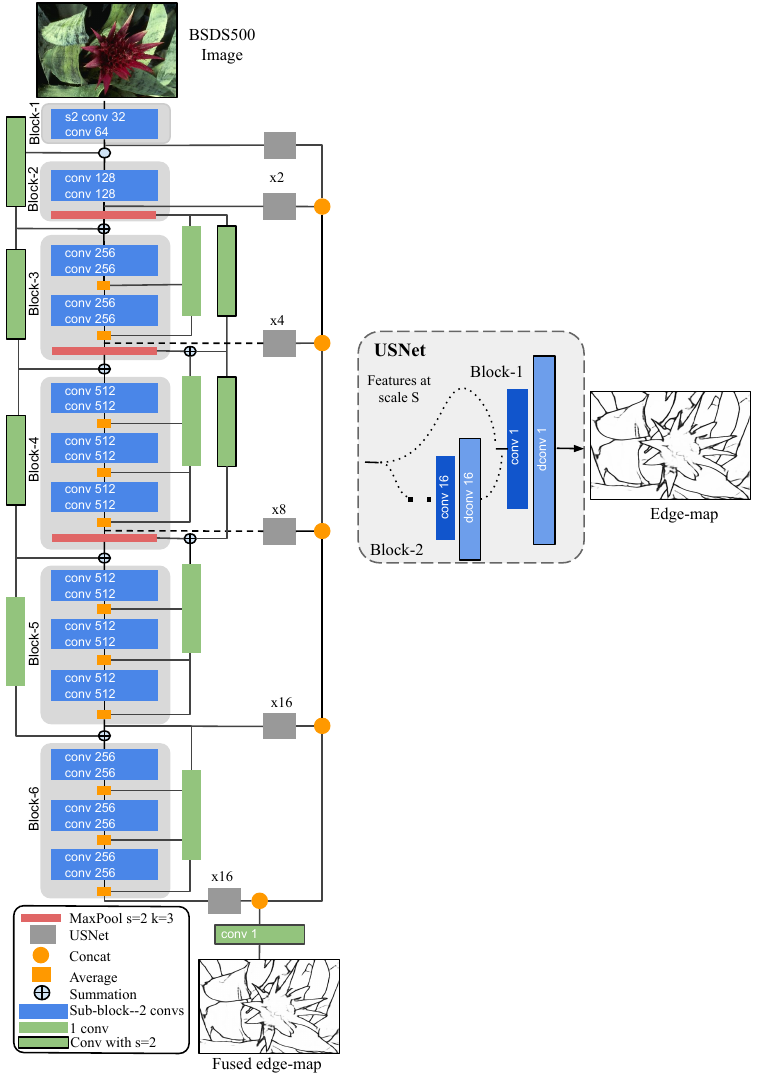}\\
\caption{Flowchart of proposed architecture (DexiNed). It consists of two building blocks a \textit{Dense Extreme Inception Network} and an upsampling Net (\textit{USNet}).}
\label{fig:arch}
\end{figure}



\section{Proposed Approach}
\label{sec:pa}

This section presents the proposed architecture and the loss function used to train the model proposed in the manuscript.

\subsection{Dense Extreme Inception Network for Edge Detection}
\label{sec:dexined}

DexiNed is designed to allow an end-to-end training without the need to weight initialization from pre-trained object detection models, like in most of DL based edge detectors. In our previous work \cite{soria2018ms-hed} we observed that the edge-features computed in shallow layers are often lost in the deeper layers. This inspired us to design an architecture similar to Xception \cite{chollet2017xception} but with two parallel skip-connections that materialize on all the edge information computed across different layers. DexiNed can be interpreted as a collection of two sub-networks (see Fig. \ref{fig:arch}): the dense extreme inception network (Dexi) and the upsampling network (USNet). Dexi receives an RGB image as input processing it in different blocks, whose feature-maps are fed to USNet.

\subsection{Dexi}
The Dexi architecture contains six blocks acting similar to an encoder. Each block is a collection of smaller sub-blocks with a group of convolutional layers. Skip-connections couple the blocks as well as their sub-blocks (depicted in light gray and blue rectangular shapes in Fig. \ref{fig:arch}). The feature-maps generated at each of the blocks are fed to a separate USNet to create intermediate edge-maps. These intermediate edge-maps are concatenated to form a stack of learned filters. At the very end of the network, these features are fused to generate a single edge-map.

Each sub-block (blue rectangles in Fig. \ref{fig:arch}) constitutes two convolutional layers (the number of kernels is specified at the right side of blue rectangles). All kernels are of size $3\times3$. In the very first block, convolutions are with a stride of $2$, hence $s2$ in its name. Each convolutional layer is followed by batch normalization and a rectified linear unit (ReLU). From the Block 3 (light grey rectangles) the last convolutional, of the last sub-block, does not contain ReLU function. Rectangles in red are max-polling operators with a $3\times3$ kernel size and stride of $2$. 
  
  
  


The multitude of convolutional operations over the depth of processing causes important edge features to vanish \cite{shen2015deepcontour,soria2018ms-hed}. To avoid this issue we introduce parallel skip-connections, inspired by \cite{he2016resnet} and \cite{zhang2018densenet}. From the third block (Block-3) forward, the output of each sub-block is averaged with another skip-connection termed \textit{second skip-connections---SSC} (green rectangles on the right side of Fig. \ref{fig:arch}). After the max-pooling operation, these SSC average the output of connected sub-blocks prior to summation with the \textit{first skip-connection---FSC},green rectangles on the left side of Fig. \ref{fig:arch}. In parallel to this, the output of max-pooling layers is directly fed to subsequent sub-blocks. For instance, the sub-blocks of block-3 receive input from the first max-pooling; and the sub-blocks of block-4 receive input with a summation of the first and second max-pooling.

\subsection{USNet}
\label{sec:pa-upsampling}

The upsampling network (USNet) is a conditional stack of two blocks. Each one of then consists of a sequence of one convolutional and deconvolutional layer that up-sample features on each pass. The block-2 gets activated only to scale the input feature-maps from the Dexi network. This block is iterated until the feature-map reaches a scale twice the size of the GT. Once this condition is met, the feature-map is fed to the block-1. The block-1 processes the input with a kernel of size $1\times1$, followed by a ReLU activation function. Next, it performs a transpose convolution (deconvolution) with a kernel of size $s \times s$, where $s$ is the input feature-map scale level. With the last deconvolution of block-1, the feature-map reaches the same size as the GT. The last convolutional layer does not have an activation function.

We considered three strategies for upsampling blocks: bi-linear interpolation, sub-pixel convolution and transpose convolution. This is an influential factor in generating thin edges, a desired feature that, for example, enhances the visualization of edge-maps. We considered this point in the design of DexiNed and a detailed evaluation of it is presented in Sec. \ref{sec:res}.


\subsection{Loss Functions}
\label{sub:pa-loss}

The DexiNed can be formalized as a regression mapping function $\eth(:)$; in other words, $\hat{Y}$ = $\eth(x,y)$, where the input image is \(x \in \mathbb{R}^{m\times d\times c}\) and \(y \in \left\{0,1\right\}^{m\times d\times1}\) is the corresponding ground truth edge-map, $m$, $d$, and $c$ are image sizes---$c$ is set to $3$ due to the RGB channels used in the model's training. The output $\hat{Y}$ corresponds to a set of predicted edge-maps, $\hat{Y} = [\hat{y}_{1},\hat{y}_{2},...,\hat{y}_{N}]$, $\hat{y_n}$ is the edge-map predicted from the $n$ DexiNed predictions, $N$ is the number of outputs from DexiNed (rectangles in gray, Fig. \ref{fig:arch}) and the last fused edge-map. $\hat{y}_{N}$ corresponds to the result of the last convolutional layer, which fuse the concatenation of all $\hat{y}_{n}$ (rectangles in gray), when the weight is initialized by $1/(N-1)$. Note that $\hat{y}_{N}$ has the same size as $y$. The loss function in our preliminary work \cite{soria2020dexined} was considered from HED \cite{xie2017hed} (weighted cross-entropy). In the current work loss functions from HED, RCF and BDCN have been evaluated; after such evaluation the BDCN loss function, $\mathcal{L}$, has been selected with a little modification as depicted below:

\begin{equation}
\centering
\begin{split}
w_{(y>0)} =1*\frac{y_-}{y_+ + y_-}; 
w_{(y<=0)} =1.1*\frac{y_+}{y_+ + y_-}, \\
l=mean(l^{I\times J}), l_{(i,j)}=[l_1,...,l_N],\\
l_n=-w[y*\log\theta(\hat{y}_n) + (1-y)*\log(1-\theta(\hat{y}_n)) ]
\end{split}
\label{eq:sin-loss}
\end{equation}

\noindent then,

\begin{equation}
\centering
\mathcal{L}= \sum_{n=1}^{N} \lambda_n *l_n
\label{eq:sum-loss}
\end{equation}


\noindent where $w$ is the weight of the cross-entropy loss, $y_-$  and $y_+$ denote negative and positive edge samples in the given GT, respectively. $\theta$ is the sigmoid function. The $\lambda$ is a set of hyper-parameters used to balance the number of positive and negative samples. Those values will be given in the implementation details section. 

\section{Datasets}
\label{sec:data}

This section details the datasets used for training and evaluation the proposed approach as well as to compare it with state-of-the-art approaches.

\subsection{Training with BIPED Dataset}
\label{sub:BIPED}

\begin{figure*}[ht] \small
    \centering
    \begin{tabular}{ccc}
        \includegraphics[width=0.28\linewidth]{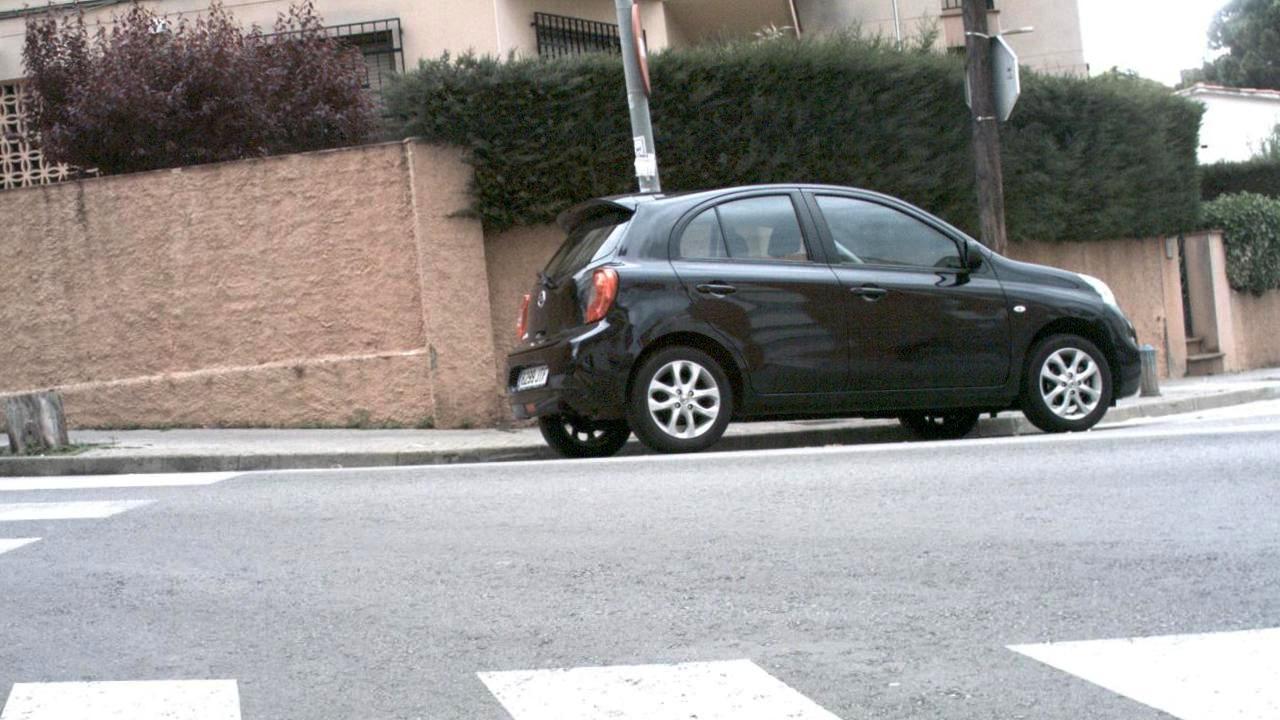}& \includegraphics[width=0.28\linewidth]{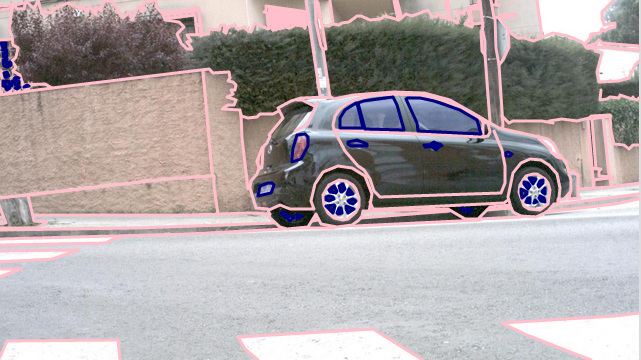}&
        \includegraphics[width=0.28\linewidth]{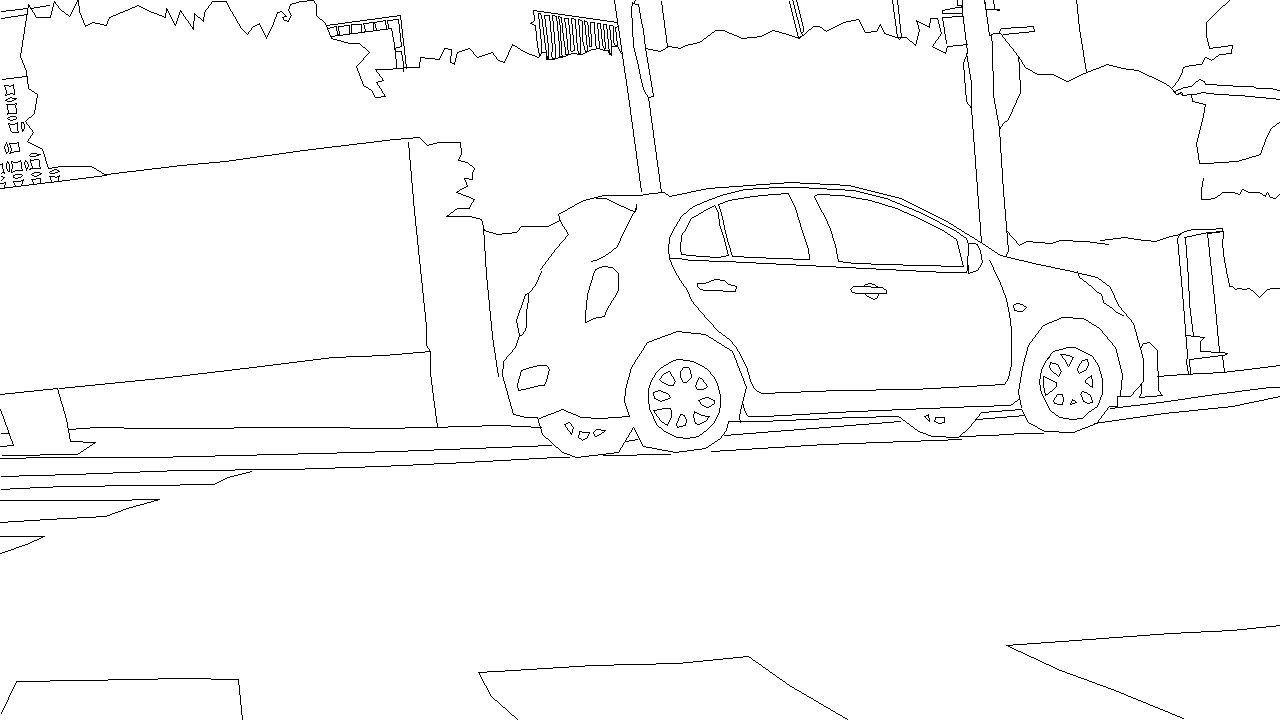}\\
        $(a)$ Original RGB image&
        $(b)$ Super imposed edge-map&
        $(c)$ Annotated edge-map\\
    \end{tabular}
    \caption{A sample image from the BIPED dataset.}
    \label{fig:BIPED_sample}
\end{figure*}

The second contribution of the current work is the update of the original benchmark dataset, referred to as Barcelona Images for Perceptual Edge Detection (BIPED), which has been initially presented in our previous work \cite{soria2020dexined}. The update presented in the current work is detailed below. All the images contained in the seminal work have been processed again. The BIPED dataset contains 250 real-world images, of $1280 \times 720$ pixels, of urban environments. Ground truth (GT) edge-maps have been generated using the crowdsourcing tool Labelbox\footnote{https://labelbox.com/}. Figure \ref{fig:BIPED_sample} presents an illustration from this dataset along with its annotated edges. The pink lines in Fig. \ref{fig:BIPED_sample}$(b)$, which are polylines, correspond to edges used to draw open contours in the scene; on the contrary, blue lines, which are polygonal lines, correspond to closed contours in scene. Figure \ref{fig:BIPED_sample}($c$) shows the edge-map corresponding to the Fig. \ref{fig:BIPED_sample}($a$). The annotation process has been applied over the whole dataset. 
To generate high precision edge-maps, each image was processed following four steps:
\begin{figure}
\begin{center}
        \includegraphics[width=0.95\textwidth]{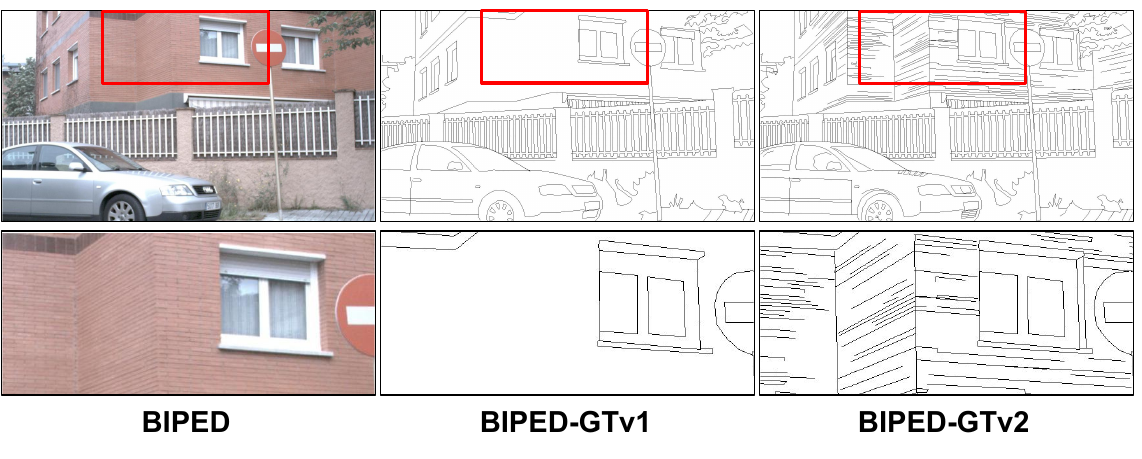} 
\end{center}
   \caption{Difference of BIPED GT previous annotation and the proposed version, BIPEd-GTv2.}
\label{fig:bipedgtv}
\end{figure}

\begin{enumerate}
    \item Computer vision experts annotate all the edges on each image, just one annotation per image.
    \item The administrator (also a computer vision expert) reviews the output of the previous step.
    \item The obtained GTs, are used to train the HED model \cite{xie2017hed} and validate the sanity of predicted edges.
    \item Considering the results of HED, the administrator crosscheck the entire dataset correcting mistakes and adding new annotations. This version was presented in \cite{soria2020dexined}.
    \item With the updated version of LabelBox, improvements in Zoom tool make us to see additional details that were not considered in the previous version, due to the lack of deep appreciation on BIPED images. This make us to add more annotation in almost the whole images. The differences between initial annotations (BIPED-GTv1) and current one (BIPED-GTv2) can be appreciated in Fig. \ref{fig:bipedgtv}; new annotation tools allow to draw very tinny edges from the given images.
\end{enumerate}

The administrator remains the same person throughout the whole process to ensure the same set of criteria is applied to all annotated images. The BIPED is a suitable dataset to benchmark results of edge detection algorithms. To this end, we have released it publicly for the benefit of the research community. The contribution of BIPED and DexiNed can be appreciated on recent works \cite{naumann2020DXNcite1, tian2020DXNcite2}, just to mention a few.

\subsection{Testing with Benchmark Datasets}
\label{sub:test-data}

A comprehensive qualitative evaluation of the proposed model has been conducted on the datasets most widely used in the literature for edge, contour, and boundary detection, namely MDBD \cite{mely2016multicue}, BSDS500 \cite{arbelaez2011bsds500}, BSDS300 \cite{martin2004mBSDS300ext}, NYUD \cite{silberman2012NYUD}, PASCAL-CONTEXT \cite{mottaghi2014PASCALcontext} (refereed hereinafter to as PASCAL), CID \cite{grigorescu2003cid}, DCD \cite{lips2019DCD} and our proposed BIPED. Within this list, MDBD is the most relevant dataset to the presented model, given its annotations correspond to true edges. The other datasets are more appropriate for the task of boundary and contour detections. In spite of that, results on all these datasets are presented to show the effectiveness of our approach and have a better comparison to state of the art.

\textit{MDBD:} The Multicue  Dataset for Boundary Detection was collected for psychophysical studies of object boundary perception in complex images. Multiple cues such as luminance, color, motion and the binocular disparity have been considered in its design \cite{mely2016multicue}. The MDBD is composed of short binocular video sequences of natural scenes (containing 10 frames per scene). Hundred high definition images of size $1280\times720$ were extracted from these videos. Each image has been annotated by several participants (six times for edge detection and five times for boundary detection). The proposed DexiNed architecture has been evaluated with the edge annotations of this dataset. Usually, in this dataset, $80\%$ of images are used for training, while the remaining are used to evaluate the learning-based algorithm. Following this, we randomly selected $20\%$ of images to evaluate the performance of DexiNed. The models considered for quantitative evaluations have been trained again for a fair comparison with the $80\%$ selected in the current work.

\textit{CID:} The Contour Image Database is a set of $40$ gray-scale images with their corresponding ground truth contours \cite{grigorescu2003cid}. The size of the images in this dataset is $512\times512$. The main limitation of this dataset is related to the small number of annotated images. The proposed DexiNed architecture has been evaluated on the entire dataset, similarly to previous works in the literature (e.g., \cite{akbarinia2018feedback, tang2007mBI3}). This dataset is difficult for DL-based approaches due to their gray-scale nature and missed annotations in some of the provided images.

\begin{figure*}
\begin{center}
\begin{tabular}{ccc}
        \includegraphics[width=0.47\textwidth]{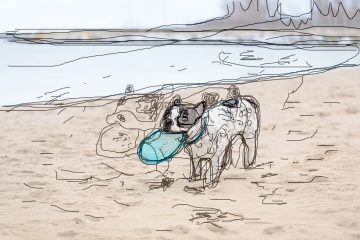} &
        \includegraphics[width=0.47\textwidth]{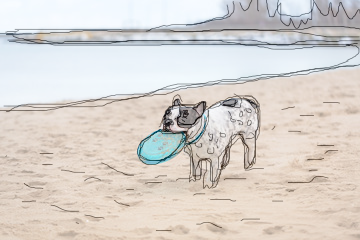} \\
        (\textit{provided annotations (5)}) & (\textit{resulting annotations (3)})\\        
\end{tabular}   
\end{center}
   \caption{($left$) A DCD test image with the provided annotations \cite{lips2019DCD}. ($right$) Contours after removing wrong annotations.}
\label{fig:dcd-errors}
\end{figure*}

\textit{DCD:} Dataset of Contour Drawings is collected with the aim to generate contour drawings (boundary-like drawings that capture the outline of a given visual scene) \cite{lips2019DCD}. The DCD contains 1000 images, from that set 100 images are selected for testing. Ground truth annotations were collected by using a game application. For each image several annotations were collected; then, after an exhaustive evaluation, just five contours per each image were kept as ground truths. In the current work, after checking the provided ground truth, several wrong annotations have been found; Figure \ref{fig:dcd-errors}($left$) shows an illustration where it can be easily appreciated the wrong contours provided by \cite{lips2019DCD}. Hence, in order to perform a fair evaluation of trained architectures the annotations of the 100 testing images have been carefully checked by removing wrong GTs. Hence, in the resulting set, which still has 100 images, each image contains between two and five annotations. Figure \ref{fig:dcd-errors}($right$) shows the results after detecting and removing wrong contours provided by \cite{lips2019DCD}.

\textit{BSDS500:} Berkeley Segmentation DataSet (BSDS), the first version has been published in 2001 \cite{martin2004mBSDS300ext} and consists of 300 images split up into 200 for training and 100 for validation, termed BSDS300; the last version \cite{arbelaez2011bsds500}, adds 200 new images for the testing. Every image in BSDS is annotated at least by $5$ subjects, this dataset contains images of $481\times321$. This dataset is mainly intended for image segmentation and boundary detection, therefore, as it will be illustrated in the next sections, for the edge detection purpose some images are not well annotated. Generally, to evaluate the performance of a DL model in BSDS500, the new 200 images are used for testing while the BSDS300 is used for the network training purpose. As DexiNed is trained only on BIPED dataset, the qualitative evaluation depicted in Section \ref{sec:res} split up the results in two parts: BSDS300 and BSDS500. The images considered for qualitative comparison are taken from the test part of BSDS500 and validation part from BSDS300.

 
\textit{NYUD:} New York University Dataset is a set of 1449 RGBD images from 464 indoor scenarios, intended for segmentation purposes. This dataset is split up into three subsets---i.e., training, validation and testing sets. The testing set contains 654 images, while the remaining images are used for training and validation purposes. In the current work, just the testing set has been selected for evaluating the proposed model, since DexiNed has been trained just with BIPED. Although most of the images in NYUD are fully annotated for segmentation, there are a few of them with poor annotations. This fact (missing edges in some images) affects the quality of DL based edge detection approaches.

\textit{PASCAL:} The PASCAL \cite{mottaghi2014PASCALcontext} is a popular dataset used for segmentation with a wide variety of object categories. Currently, most of the major DL methods for edge detection use PASCAL for training and testing (e.g., \cite{xie2017hed}, \cite{liu2019RCFext}), due to its ground-truths correspond to scenes different to the ones depicted in BSDS. This dataset contains 11530 annotated images, however, just around $5\%$ of randomly selected images (505) have been considered for testing DexiNed. Although the images in PASCAL have more diverse labeled data, most of the images are annotated only for a couple of objects even though the scene has a vast number of features.

\section{Experimental Results}
\label{sec:res}

This section firstly describes the metrics used for the evaluations; then, details about the implementation and settings of the proposed approach are provided. Finally, a large set of experimental results is presented together with comparisons with state-of-the-art approaches. Again, most of the non-edge detection datasets on the literature are used just for qualitative evaluation and, for the edge detection, BIPED and MDBD approaches, are used for quantitative evaluations and comparisons with DexiNed.

\subsection{Evaluation Metrics}
\label{sub:metrics}

Edge detection algorithms can be evaluated following two approaches. \textit{Indirectly} through their impact on other computer vision tasks \cite{ziou1998edgeOverview}. \textit{Directly} in comparison to human drawn edges. In this work, we opted for the latter, which is a common practice in the evaluation of benchmark datasets.
These evaluation metrics are as follow:
\begin{enumerate}
    \item Optimal Dataset Scale (ODS) computed by using a global threshold for the entire dataset;
    \item Optimal Image Scale (OIS) computed by using a different threshold on every image;
    \item Average Precision (AP). The F-measure---$F=\frac{2\times Precision \times Recall}{Precision + Recall}$.
\end{enumerate}

\subsection{Implementation Notes}
\label{sub:impl-notes}

The proposed DexiNed architecture has been trained from scratch without relying on pre-trained weights. This is a unique feature of the proposed model. Most state-of-the-art networks depend on pre-trained weights of the ImageNet dataset. On average, DexiNed converges after 9 epochs (15 epochs in \cite{soria2020dexined}) with a batch size of 8 using Adam optimizer and learning rate of $10^{-4}$---decreasing at 10 and 15 epochs by a factor of 0.1, the weight decay considered for the training was $10^{-8}$. The training procedure takes around 1 day in a TITAN X GPU input with color images of size 352$\times$352, 480$\times$480 for MDBD. The weights for fusion layer are initialized as: $\frac{1}{N-1}$ (see Section \ref{sub:pa-loss} for details on $N$). After a hyperparameter search to optimize DexiNed, the best performance was obtained using kernels of size 3$\times$3, 1$\times$1 and s$\times$s on the different conv and deconv layers, with Xavier initializer \cite{glorot2010xavint} in Dexi and normal distribution in the last conv and deconv layers of USNet. The $\lambda$ of the loss function is set tp [0.7, 0.7, 1.1, 1.1, 0.3, 0.3, 1.3].

We randomly selected 200 images of BIPED to train and validate DexiNed. The remaining 50 images were used for testing. In order to increase the number of training images, a data augmentation process has been performed as follow: i) given the high resolution nature of BIPED images, each image is split in half along its width; ii) similarly to HED, each of the resulting images is rotated by 15 different angles and crop by the inner axis oriented rectangle; iii) images are horizontally flipped; and finally iv) two gamma corrections have been applied (0.3030, 0.6060). This augmentation process resulted in 288 images per each of the given images. 

 
\subsection{Architecture Setup}
\label{sub:experim}

This section presents evaluations on different DexiNed configurations. The first sub-section presents details on the upsampling methods and the merging process selected for estimating the edge-maps. In the second sub-section, evaluations on the advantage of using skip connections are presented.


\subsubsection{Upsampling Methods and DexiNed Predictions}

Regarding the selection of the best upsampling method, in our preliminary work \cite{soria2020dexined} three approaches were evaluated. In that evaluation it is shown that the upsampling performed by the transpose convolution with trainable kernels gives the best results. Hence, in the current work it is also selected as upsampling method. On the other hand, regarding the strategy used to merge the different outputs---Output1, Output2, Output3, Output4, Output5, Output6, DexiNed-f, DexiNed-a---the DexiNed architecture (Fig. \ref{fig:arch}) has been empirically evaluated as performed in \cite{soria2020dexined} showing that the best results are obtained from DexiNed-f and DexiNed-a. DexiNed-f corresponds to the result obtained by the fusion process at the end of the DexiNed architecture (see Fig. \ref{fig:arch}), while DexiNed-a corresponds to the edge-maps obtained from the average of all DexiNed predictions (DexiNed-f included). It is shown in \cite{soria2020dexined} that both merging strategies reach similar quantitative results; hence hereinafter results from both merging strategies are presented; in some cases, due to space limitations, just results from DexiNed-f are depicted, in those cases results are named as DexiNed.

\subsubsection{Ablation Study}\label{sec:skip}

\begin{figure}
    \centering
    \includegraphics[width=0.98\textwidth]{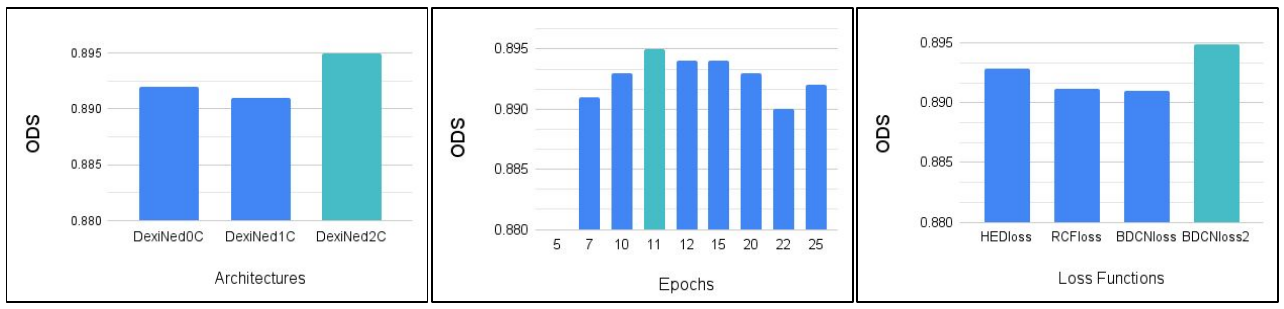}

    \caption{ $(left)$ Evaluation of DexiNed architecture with (DexiNed1C: 1 connection; DexiNed2C: 2 connections) and without (DexiNed0C) skip-connections. $(middle)$ DexiNed performance evolution during training. $(right)$ Evaluation with different loss functions.}
    \label{fig:ablation}
\end{figure}

This subsection presents a quantitative study of the critical DexiNed parts and settings. As our model is composed with two types of skip connections (rectangles in green in the left and right sides in Fig. \ref{fig:arch}) a study of different number of skip-connections is performed. It is presented in Fig. \ref{fig:ablation}$(left)$---$DexiNed0C$ does not use skip-connections, $DexiNed1C$ uses just one skip connection (left side in Fig. \ref{fig:arch}); $DexiNed2C$ uses the two connections. It is clear that the usage of two skip-connections (DexiNed2C) improves the performance of the proposed architecture.

The proposed architecture has been also trained with different loss functions to evaluate its performance. Loss functions from the following approaches have been considered: HED \cite{xie2017hed}, RCF \cite{liu2019RCFext}, BDCN \cite{he2020bdcnPAMI}, and BDCNloss2 (it is a slightly modified function from BDCN \cite{he2020bdcnPAMI}). As illustrates in the Fig. \ref{fig:ablation}$(right)$ the modified BDCN loss function outperforms its counterparts with almost $1\%$. Finally, to choose the best DexiNed performance, we have evaluated its prediction in different epochs, till 25 epochs. As it can be appreciated in Fig. \ref{fig:ablation}$(middle)$ the best performance is reached when 11 epochs are considered. In the following sections of this manuscript, the comparisons of DexiNed performance on the different edge detection datasets correspond to the architecture DexiNEd2C, with 11 epochs, and using BDCNloss2 function, termed just as DexiNed.

        


\subsection{Quantitative Comparison}
\label{sub:quant}

\begin{table}[b]
\scriptsize
\caption{Quantitative results: The performance of DexiNed model and BIPED dataset are compared to the last DL based models and the edge detection based dataset, MDBD\cite{mely2016multicue}.}
\label{tab:alledge2}
\centering
\begin{tabular}{cccccccccc}
\hline
Methods& Trained on& Tested on&ODS&OIS&AP&Tested on&ODS&OIS&AP\\
 \hline\hline
 RCF\cite{liu2019RCFext} (2019)
 &\multirow{5}{*}{\rotatebox[origin=c]{90}{MDBD \cite{mely2016multicue}}} &\multirow{5}{*}{\rotatebox[origin=c]{90}{MDBD \cite{mely2016multicue}}}&.879&.888&.926
 &\multirow{5}{*}{\rotatebox[origin=c]{90}{BIPED}}&.8140&.828&.871\\
 BDCN\cite{he2020bdcnPAMI} (2020)&&&.887&.891&.793
 &&\textbf{.854}&\textbf{.863}&.768\\
 CATS\cite{huan2021cats} (2021)&&&.891&.899&.809&&.813&.831&.791\\
DexiNed-f (Ours)&&&.891&.896&.930&&.787&.807&.844\\
DexiNed-a (Ours)&&&\textbf{.894}&\textbf{.902}&\textbf{.951}&&.789&.813&\textbf{.875}\\
\hline
 RCF\cite{liu2019RCFext} (2019)
 &\multirow{5}{*}{\rotatebox[origin=c]{90}{BIPED}} &\multirow{5}{*}{\rotatebox[origin=c]{90}{BIPED}}&.849&.861&.906
 &\multirow{5}{*}{\rotatebox[origin=c]{90}{MDBD \cite{mely2016multicue}}}&.839&.854&.865\\
 BDCN\cite{he2020bdcnPAMI} (2020)&&&.890&.899&.934
 &&.855&.864&.692\\
 CATS\cite{huan2021cats} (2021)&&&.887&.892&.817&&.837&.840&.496\\
DexiNed-f (Ours)&&&\textbf{.895}&\textbf{.900}&.927&&\textbf{.863}&.871&.867\\
DexiNed-a (Ours)&&&.893&.897&\textbf{.940}&&.862&\textbf{.874}&\textbf{.919}\\
 \hline

\end{tabular}

\end{table}

This section presents comparisons of DexiNed with the state-of-the-art approaches on edge detection datasets. Additionally, a comparison of the  generalization capability in two datasets (i.e., MDBD and BIPED) is studied; in other words, a study that shows results of generalization from one dataset to another dataset is presented. As introduced in Sec. \ref{sec:intro}, BSDS \cite{arbelaez2011bsds500} is not considered since this dataset cannot generalize results on edge domain. The state-of-the-art approaches for edge, contour, and boundary detection \cite{Gong2018contourOverview} have been selected for comparison with DexiNed; these approaches are the following: RCF \cite{liu2019RCFext}, BDCN \cite{he2020bdcnPAMI}, and CATS \cite{huan2021cats}. In order to perform a fair comparison, these approaches have been trained on two datasets intended for edge detection---the MDBD and our BIPED datasets. It should be noticed that the same augmentation processes has been applied in all the cases; additionally, in the case of MDBD, all models have been trained with the same training set, instead of randomly selecting images from the MDBD dataset, as performed in most of the publications.

Table \ref{tab:alledge2} shows different evaluations for DexiNed (DexiNed-f and DexiNed-a) and the approaches from the state of the art mentioned above. Results from both scenarios are presented---trained and tested on the same dataset and cross-evaluations (i.e., trained in a dataset and evaluated in the other dataset). It can be appreciated that DexiNed reaches the best performance (in all three metrics ODS, OIS, and AP) when trained and tested in the same dataset. Furthermore, DexiNed reaches also the best result if trained in BIPED but evaluated in the MDBD dataset. On the other hand, regarding the dataset generalization capability, considering ODS evaluation metric, we know that the best performance is reached when trained and evaluated in the same dataset. Hence, we propose to analyze the loss in performance when evaluated in another dataset different to the one used for training. In other words, what is the performance with respect to the values reached when trained and evaluated in the same dataset. This loss in performance is smaller if the different approaches are trained in BIPED, on average just 3,74\% of decrease in performance is appreciated when evaluated in MDBD; on the contrary, this loss in performance reaches on average 7,14\% if trained in MDBD and evaluated in BIPED.

\subsection{Qualitative Comparison}
\label{sub:qualr}

 	


 \begin{figure*}
 	\centering
 	\includegraphics[width=1\textwidth,height=0.75\textheight]
 	{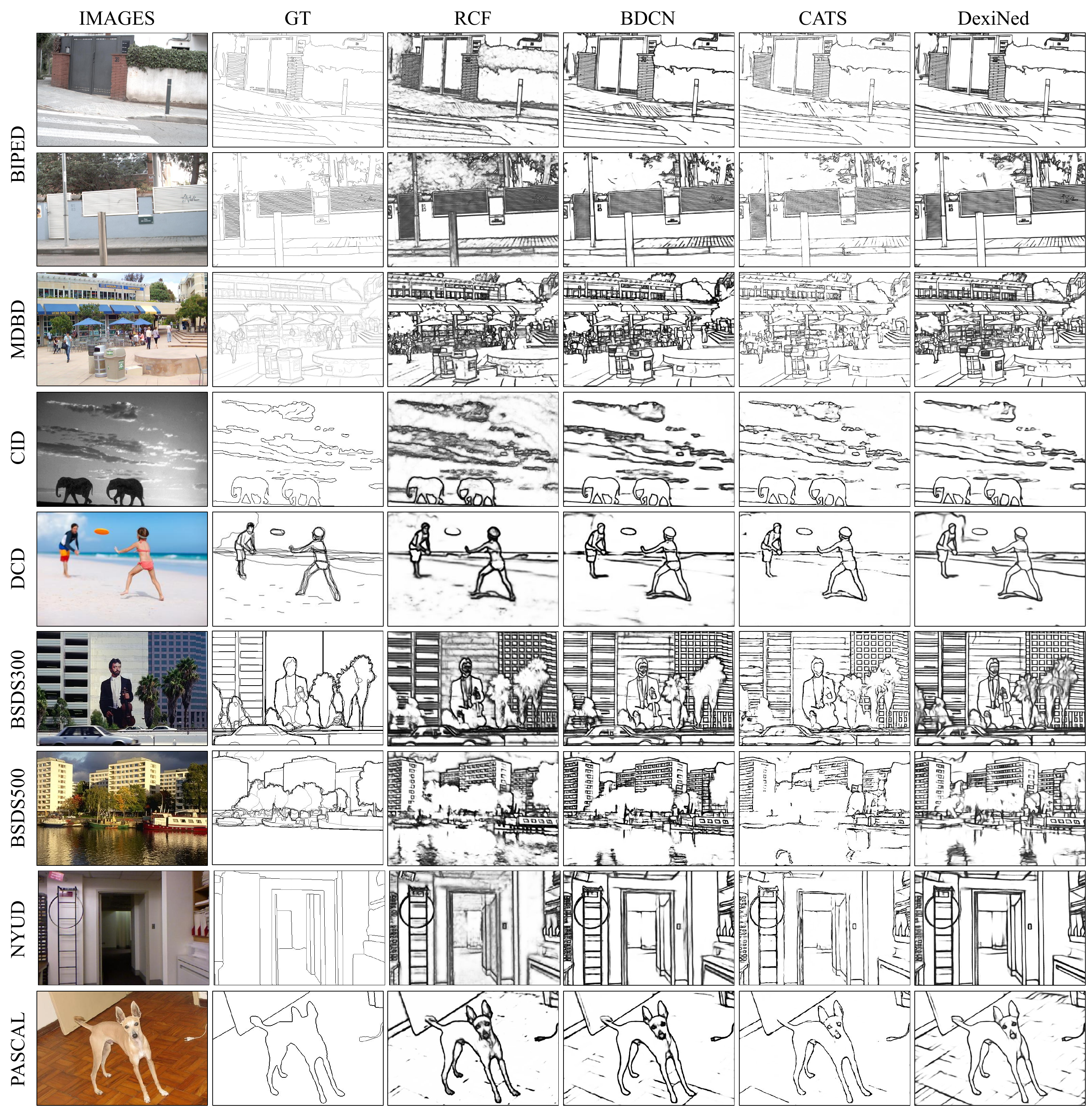}\\
 	\caption{The results of a few state-of-the-art architectures trained on the proposed BIPED dataset. Note that the ground truth of these datasets, except for BIPED and MDBD, correspond to boundary and segmentation tasks.}
 	\label{fig:quali}
 \end{figure*}

This section presents just some illustrations as qualitative comparisons of the edge-maps predicted from all the models considered during the quantitative comparison in the previous section. Note that all these models have been trained on BIPED; Figure \ref{fig:quali} shows edge-maps of the evaluated architectures (trained on BIPED) when used in the datasets detailed in Section \ref{sub:test-data}. Qualitative results, similar to those presented in Fig. \ref{fig:quali}, but trained on MDBD \cite{mely2016multicue}, are presented as supplementary material. Some comments and conclusions from the analysis of the obtained results are presented below:

\begin{itemize}
    \item \textit{BIPED:} For this dataset two illustrations are presented since this dataset is used for training the selected models considered in the comparisons. These two illustrations correspond to the best and the worst ODS results from DexiNed. As shown in Fig. \ref{fig:quali}, edge maps predicted by BDCN, CATS, and DexiNed are clean and accurate representations. Perceptually, we can say that edge maps predicted by BDCN and DexiNed contains more correctly detected edges. On the contrary, edge maps predicted by CATS are cleaner but containing less predictions. As a conclusion, while searching for a model that predicts more edges but also less artifacts, DexiNed provide us those requirements without compromise them.
    
    \item \textit{Other datasets: MDBD, CID, DCD, BSDS300, BSDS500, NUYD, and PASCAL.} As shown in Fig. \ref{fig:quali}, results in these datasets are also similar to predictions from the test set on BIPED. The CATS model shows cleaner edge-maps but with less edges than DexiNed. A more perceptual difference of this claim can be observe in the predictions of MDBD, CID, NYUD, and PASCAL. Looking at the last row in Fig. \ref{fig:quali}, which corresponds to PASCAL dataset, several edges on the floor of the scene have been detected by DexiNed, but none of the other models are able to detect them. 
    

\end{itemize}

To finish, the second version of BIPED gives the generalization robustness to the models considered for comparison. Overall, in all the datasets presented in Fig. \ref{fig:quali}, thanks to the unique characteristics on BIPED, there are most perceptual edges predicted in the images. Concerning to the  DexiNed architecture, the predictions given by this model are cleaner than from its counterparts and  without compromising true edge lost, the training procedure is simpler that in any other DL based models considered for comparison. DexiNed does not need pre-training weights, it can converge in less time than other models but still reach the state-of-the-art results.

\section{Conclusion}
\label{sec:con}

In this paper, we proposed a robust edge detection model that exhibits a great degree of generalization to new scenes. To this end, first, we presented a benchmark dataset carefully designed for the task of edge detection. Second, we designed a network with parallel skip-connections that learn edges without the need for ImageNet pre-trained weights. We demonstrated the generalization power of our approach by training a network on a single dataset and evaluating it on other benchmark datasets. Overall, our results show the possibility of training a deep learning model of edge detection from scratch in an end-to-end fashion. These findings open the opportunity to explore smaller networks for the task of edge detection by reducing the number of hyper-parameters settings. 

\section*{Acknowledgment}
This material is based upon work supported by the Air Force Office of Scientific Research under award number FA9550-22-1-0261; and partially supported by the Grant PID2021-128945NB-I00 funded by MCIN/AEI/10.13039/501100011033 and by ``ERDF A way of making Europe"; the ``CERCA Programme / Generalitat de Catalunya"; and the ESPOL project CIDIS-12-2022. A.A. was funded by Deutsche Forschungsgemeinschaft SFB/TRR135 (grant number 222641018). X.S. was funded by Ecuador government institution SENESCYT (No. 2015-AR3R7694).

\balance

\bibliographystyle{elsarticle-num}
\bibliography{reflist}

\balance{}
\bio[]{}
\textbf{Xavier Soria:} received his B.S. degree in Educational Informatics from National University of Chimborazo in 2009, Ecuador, and the Ph.D. degree in Computer Science from Autonomous University of Barcelona in 2019, Spain. Currently, He is an Adjunct Professor at National University of Chimborazo, Ecuador. His research interest includes computer vision and its applications.
\endbio
\bio[]{}
\textbf{Angel Sappa:} received the Electromechanical Engineering degree from National University of La Pampa, Argentina, in 1995, and the Ph.D. degree from the Polytechnic University of Catalonia, Spain, in 1999. Currently he is a Senior Researcher at the Computer Vision Center, Spain, and a Full Professor at the ESPOL Polytechnic University, Ecuador.
\endbio

\bio[]{}
\textbf{Patricio Humanante:} received the Systems Engineer degree from the Polytechnic School of Chimborazo, Ecuador, in 1998, and the Ph.D. degree in Training in the Knowledge Society from the University of Salamanca, Spain, in 2016. Since 1999, he is a Full Professor and Researcher at the National University of Chimborazo, Ecuador.
\endbio

\bio[]{}
\textbf{Arash Akbarinia:} received BSc. in Software Engineering from Göteborgs Universitet and MSc. in Computer Vision jointly from Université de Bourgogne, Universitat de Girona, and Heriot-Watt University. He obtained his Ph.D. from Universitat Aut\`{o}noma de Barcelona. He is currently a research scientist at Justus-Liebig-Universität Giessen. His research interests include artificial and biological vision.
\endbio


\end{document}